\documentclass[]{article}
\usepackage[left=2.5cm, right=2.5cm]{geometry}
\usepackage[utf8]{inputenc} 
\usepackage[T1]{fontenc}    
\usepackage[hidelinks]{hyperref}       
\usepackage{url}            
\usepackage{booktabs}       
\usepackage{amsfonts}       
\usepackage{nicefrac}       
\usepackage{microtype}      
\usepackage{xcolor}
\usepackage{amsmath,amssymb}
\usepackage{subcaption}
\usepackage{bm}
\usepackage{graphicx}
\usepackage{natbib}

\newcommand{\EE}{\mathcal{E}}
\newcommand{\GG}{\mathcal{G}} 
\newcommand{\VV}{\mathcal{V}} 
\newcommand{\NN}{\mathcal{N}} 
\newcommand{\R}{\mathbb{R}}
\newcommand{\C}{\mathbb{C}}
\newcommand{\E}{\mathbb{E}}

\newcommand{\vv}{\mathbf{v}}
\newcommand{\ee}{\mathbf{e}}
\newcommand{\xx}{\mathbf{x}}
\newcommand{\uu}{\mathbf{u}}
\newcommand{\zz}{\mathbf{z}}

\newcommand{\ie}{\emph{i.e.}}
\newcommand{\eg}{\emph{e.g.}}

\title{Beyond permutation equivariance in graph networks}

\author{%
  Emma Slade, Francesco Farina\\
  {\small GSK.ai}\\
  {\small GlaxoSmithKline}\\
  {\scriptsize\texttt{\{emma.x.slade,francesco.x.farina\}@gsk.com}}
}

\date{}

\begin{document}

\maketitle

\begin{abstract}
  
  In this draft paper, we introduce a novel architecture for graph networks which is equivariant to the Euclidean group in $n$-dimensions. The model is designed to work with graph networks in their general form and can be shown to include particular variants as special cases.
  Thanks to its equivariance properties, we expect the proposed model to be more data efficient with respect to classical graph architectures and also intrinsically equipped with a better inductive bias. We defer investigating this matter to future work.
  \end{abstract}

\section{Introduction}

Symmetries exist throughout nature. All the fundamental laws of physics are built upon the framework of symmetries, from the gauge groups describing the Standard Model of particle physics, to Einstein's theories of general and special relativity. Once we understand the symmetry of our system, we can make powerful predictions. A notable example is that of Gell-Mann's eightfold-way~\citep{osti_4008239}, built upon the symmetries observed in hadrons, that led to his prediction of the $\Omega^-$ baryon, which was subsequently observed 3 years later~\citep{1964PhRvL..12..204B}.

The mathematical framework used to describe symmetries, group theory, has recently been incorporated into many different deep learning models, with focus on differing symmetry groups.
Group equivariant convolutional networks are obtained in~\citep{cohen2016group} by constructing representations that have the
structure of a linear $G$-space, while a general self-attention formulation to impose group equivariance to arbitrary symmetry groups in convolutional networks is provided in~\citep{romero2021group}.
In~\citep{mattheakis2019physical}, physical symmetries are embedded in neural networks via embedding physical constraints in the structure of the network, equivariance to the Lorentz group is achieved in~\citep{bogatskiy2020lorentz} and in~\citep{barenboim2021symmetry} the ability of neural networks to discover and learn symmetries in data is explored. 

Graph networks are designed to learn from graph-structured data and are by construction permutation invariant with respect to the input.
They have been originally proposed in~\citep{gori2005new,scarselli2008graph} but received great attention only recently (see, \eg,~\citep{battaglia2018relational,hamilton2020graph,wu2020comprehensive} for a comprehensive overview). Graph networks find application in a broad range of problems like learning the dynamics of complex physical systems~\citep{sanchez2020learning, pfaff2021learning}, learning causal and relational graphs~\citep{kipf2018neural, li2020causal}, discovering symbolic models~\citep{cranmer2020discovering}, as well as seminal work in the fields of quantum chemistry~\citep{gilmer2017neural} and drug discovery~\citep{STOKES2020688}.

By construction, graph networks are permutation invariant, which for appropriately structured datatypes, such as $n$-body systems, can greatly improve performance. It is therefore natural to consider whether equivariance under more general groups may be useful for graphical models.
The role of equivariance in graph networks has only very recently been explored; in~\citep{tang2020towards} a model is presented which is equivariant on the scale of the graph itself, whilst~\citep{thiede2021autobahn} works with the automorphism group of subgraphs. 
There has been particular interest in the Euclidean group with results for the subgroups $SE(3)$ and $E(3)$ obtained in~\citep{1802-08219,NEURIPS2020_15231a7c, kohler2020equivariant, finzi2020generalizing, batzner2021se3equivariant, Yang_2020_CVPR}
and initial results for $E(n)$ in (message passing) graph convolutional networks in \citep{satorras2021en} and \citep{horie2021isometric}. In particular, the work in~\citep{satorras2021en} is built on theoretical arguments that are similar to the one we use in this paper.\footnote{The work in~\citep{satorras2021en} has been conducted in parallel and independently from us. Nevertheless, our architecture is more general, allowing for generic graph networks. The one presented in~\citep{satorras2021en} focuses on a message passing scheme for a graph convolutional network - providing extensive insights and experiments - and can be obtained as a particular case of the one we introduce here.}

In this work we take a step towards going beyond permutation equivariance in graph neural networks by defining a novel graph architecture with modified update rules that make it equivariant to Euclidean transformations of the node coordinates.
In this way, a network is able to filter out many copies of the same input simply rotated or moved and should learn more efficiently than one which considers the inputs as distinct. In other words, a single sample contains the same information as many copies of it obtained by rotating and translating it. 

This draft is meant to be a preliminary work outlining this architecture and its potential. We defer further extensions, discussions and experiments to future works.

\section{Group theory}
Symmetries of physical systems can be generally classified as continuous or discrete, with group theory providing the mathematical formulation for symmetries. Symmetry operations are represented by individual group elements and we therefore have continuous and discrete groups. Discrete groups may have infinite or finite numbers of elements, whereas continuous symmetries are described by Lie groups which are smooth manifolds with group structure.
A group is, loosely speaking, a set $G$ equipped with a binary operation $\star$, which enables one to combine two group elements to form a third, whilst preserving the group axioms
\begin{align*}
\text{Associativity} &: (a \star b) \star  c = a \star (b \star c) \quad \forall a, b, c \in G \,,\\
\text{Identity}& : \exists! \, E \in G \quad\text{such that}\quad E\star a = a \star E = a \quad \forall a \in G \,,\\
\text{Closure}&:  \exists \, a  \star b \in G \quad \forall a, b \in G \,, \\
\text{Inverse}&:  \exists \, a^{-1} \in G\quad \text{such that}\quad a \star a^{-1} = a^{-1} \star a = E \quad \forall a \in G\,.
\end{align*}
An example of a group with which we are all naturally familiar and which one can simply conclude satisfies the group axioms is the set of integers $\mathbb{Z}$ under addition. In this case the group is discrete with an infinite number of elements.

Continuous symmetries in nature include well known ones such translations and rotations in isotropic systems, as well as ones which may be more obscure; the symmetry group of quantum mechanical spins $SU(2)$, the colour group of quantum chromodynamics $SU(3)$, as well as the group describing electromagnetism $U(1)$ together describe the Standard Model of particle physics. Discrete groups are important for crystalline systems, with the rotations and translations described by the space groups.

Now we are equipped with a basic understanding of groups, we can introduce the idea of equivariance.
Let $\varphi_g :X\to X$ be a transformation on $X$ for an abstract group $g \in G$. Then, the linear map $\Phi:X\to Y$ is \emph{equivariant} to  $G$ if  $\forall \varphi_g \in X, \forall g \in G,$ $\exists\varphi_g ':Y\to Y$ such that
\begin{equation*}
 \Phi(\varphi_g(\xx)) = \varphi_g'(\Phi(\xx)) \,.
\end{equation*}
When $\varphi_g '$ is the identity, we say that $\Phi$ is invariant  to $G$.
Whilst not often expressed in group theoretic notation, equivariances exist in common deep learning architectures; convolutional neural networks are equivariant under the translation group $T(n)$, whilst graph neural networks are equivariant under the symmetric group $S_n$.

\subsection{$E(n)$ equivariance}
As mentioned above, examples of continuous groups include the translation and rotation groups ($T(n)$ and $O(n)$ respectively). The semidirect product of these two groups $E(n) = T(n) \rtimes O(n)$ is known as the Euclidean group.
In order to be $E(n)$ equivariant, mathematically, we wish to find the transformations $\varphi: \E^n \to \E^n$, where $\E^n$ is a Euclidian space. Note that $\R^n$, equipped with an inner product is a Euclidean space $\E^n$, and we assume an inner product to exist without loss of generality.
Since $E(n)$ is the semidirect product of $T(n)$ and $O(n)$ we can start by considering them separately.
Under translation, one has
  \begin{equation*}
    \xx \to \xx^+= \xx + \zz, \quad \zz \in \E^n \,,
  \end{equation*}
while under rotation
  \begin{equation*}
    \xx \to \xx^+= Q\xx, \quad Q \in O(n) \,.
  \end{equation*}
Now, by using the above relations, it is easy to show that the function $\|\xx_i-\xx_j\|_2^2$ is equivariant under $E(n)$ since
\begin{subequations}\label{eq:En}
\begin{align}
  \|\xx_i - \xx_j\|_2^2\to \|\xx_i^+ - \xx_j^+\|_2^2 &= (Q(\xx_i + \zz) - Q(\xx_j + \zz) )^\top(Q(\xx_i + \zz) - Q(\xx_j + \zz) )\\
    &= (Q\xx_i - Q\xx_j  )^\top(Q\xx_i  - Q\xx_j ) \\
    &= (\xx_i - \xx_j  )^\top Q^\top Q (\xx_i  - \xx_j ) \\
    &= (\xx_i - \xx_j  )^\top (\xx_i  - \xx_j ) \,,
\end{align}
where we used the fact that $Q^\top Q=I$ for all $Q \in O(n)$.
\end{subequations}

\section{Toward equivariant graph networks}
In order to extend graph networks beyond permutation equivariance we start by introducing graph networks in their most general form (similarly to~\citep{battaglia2018relational}). Then, we build on the notion of equivariance to redefine network updates to be equivariant to any Euclidean transformation in the node coordinates.

\subsection{Graph networks}

Let a graph be defined as $\GG=(\VV,\EE)$ where $\VV=\{1,\dots, N\}$ is the set of nodes, $\EE\subseteq \VV\times\VV$ is the set of (directed) edges connecting nodes in $\VV$. We denote $\NN_i$ as the set of neighbours of node $i$, \ie, $\NN_i=\{j\mid (j, i)\in\EE\}$. Using a similar notation as in~\citep{battaglia2018relational}, we let $\vv_i\in\R^{n_v}$ represent the attributes of node $i$ for all $i\in\VV$, $\ee_{ij}\in\R^{n_e}$ represent the attributes of edge $(i,j)$ for all $(i,j)\in\EE$ and $\uu\in\R^{n_u}$ a global attribute associated to the graph\footnote{For the sake of exposition we assume node, edge and global attributes to be vectors but the discussion can be easily extended to multi-dimensional attributes.}. 
Then, a graph network can be characterised in terms of edge, node and global updates as
\begin{subequations}\label{eq:gnn}
\begin{align}
  \ee_{ij}^+ &= \phi^e\big(\ee_{ij}, \vv_i, \vv_j, \uu\big),&\forall (i,j)\in\EE\\
  \vv_i^+ &= \phi^v\big(\rho^{e\to v}\big(\{\ee_{ij}^+\}_{j\in\NN_i}\big), \vv_i, \uu\big),&\forall i\in\VV\\
  \uu^+ &= \phi^u\big(\rho^{e\to u}(\{\ee_{ij}^+\}_{(i,j)\in\EE}),\rho^{v\to u}(\{\vv_{i}^+\}_{i\in\VV}), \uu\big)
\end{align}
\end{subequations}
where $\phi^e:\R^{n_e+ 2n_v + n_u}\to \R^{n_e}$, $\phi^v:\R^{n_e+ n_v+ n_u}\to \R^{n_v}$, $\phi^u:\R^{n_e+ n_v+ n_u}\to \R^{n_u}$ are the update functions to be learned and $\rho^{e\to v}, \rho^{e\to u}, \rho^{v\to u}$ are aggregation functions reducing a set of elements to a single one via some input's permutation equivariant transformation (like mean, sum, max\dots).

While updates~\eqref{eq:gnn} are permutation equivariant by construction, a natural question that arises is whether equivariance to other groups can be achieved within the same framework.
\begin{figure}
  \centering
  \begin{subfigure}{0.24\textwidth}
    \centering
    \includegraphics[width=0.8\textwidth]{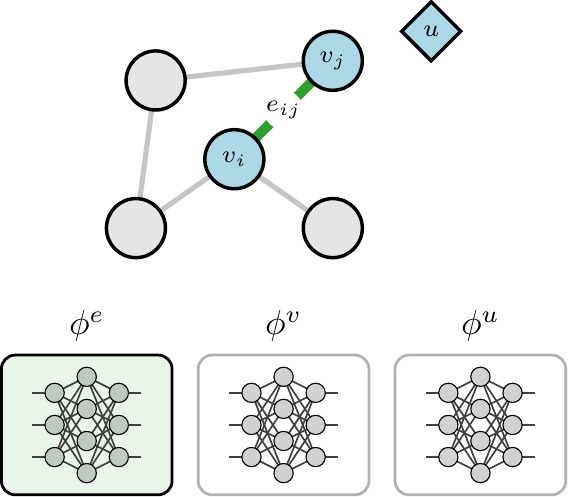}
    \caption{Edge update}
  \end{subfigure}
  \begin{subfigure}{0.24\textwidth}
    \centering
    \includegraphics[width=0.8\textwidth]{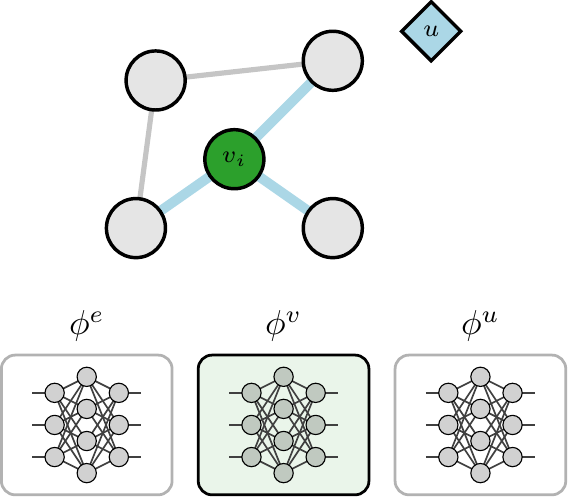}
    \caption{Node update}
  \end{subfigure}
  \begin{subfigure}{0.24\textwidth}
    \centering
    \includegraphics[width=0.8\textwidth]{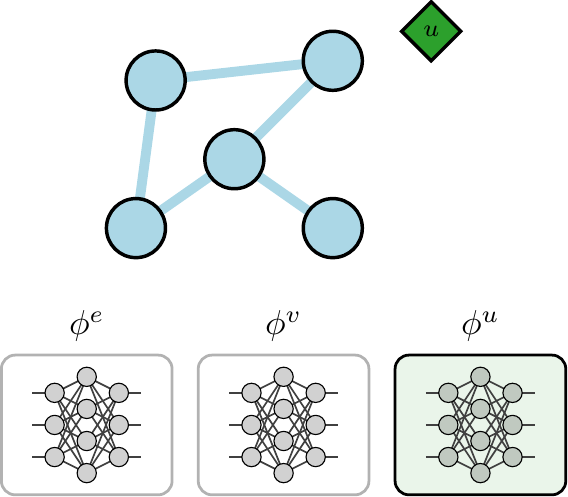}
    \caption{Coordinate update}
  \end{subfigure}
  \begin{subfigure}{0.24\textwidth}
    \centering
    \includegraphics[width=0.8\textwidth]{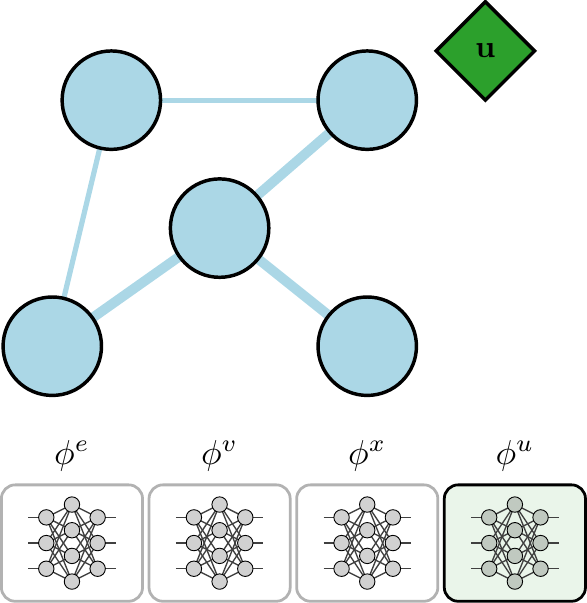}
    \caption{Global update}
  \end{subfigure}
  \caption{$E(n)$ graph network updates. In green the object being updated (along with the corresponding update function), in blue the quantities used to perform the update and in grey the unused objects.}
  \label{fig:gnns}
\end{figure}

We note that, while our discussion here is restricted to the space of reals, $\R^n$, the arguments can be extended to the space of complex numbers.
In particular, in two dimensions, $\R^2 \cong \C$, one may extend our arguments to the conformal group $\text{Conf}(\R^{2,0})$, which preserves only angles and considers the transformations $\varphi : \R^2 \to \R^2$. The conformal group in two dimensions has particularly important properties, as the transformations of the conformal group  $\text{Conf}(\R^{2,0}) \cong SO(3,1; \R)\cong SL_2(\C)$, where $SO(p,q)$ is the group of linear transformations with determinant 1 which leave a symmetric bilinear form $(p,q)$ invariant and $SL_2(\C)$ is the set of $2\times 2$ matrices over $\C$ with determinant 1. We leave further discussion of this to future work.

\subsection{Building equivariance to the Euclidean group}
The Euclidean group encapsulates symmetries which may exist in large amounts of physical data, such as an image (or representation of an image) which may have been rotated or moved with respect to the centre.
It is therefore natural to try to build a deep learning architecture which has $E(n)$ equivariance built into it; a network which is able to filter out many copies of the same image simply rotated or moved should learn more efficiently than one which considers the images to be distinct.

In order to build an $E(n)$ equivariant graph network, we must work directly with the coordinate features, $\xx_i\in\R^{n_x}$ for each node $i\in\VV$. There is no sense of equivariance of the network's nodes or edges as they are not embedded in $\E^n$ due to the lack of definition, in general, for an inner product between nodes or edges.

 If we assume that the initial node attributes $\vv_i$ contain no absolute coordinate or orientation information about the initial coordinates $\xx_i$ then a form of $E(n)$ equivariant updates is given by 
 \begin{subequations}
  \begin{align}
    \ee_{ij}^+ &= \phi^e\big(\ee_{ij}, \vv_i, \vv_j, \|\xx_i-\xx_j\|_2^2, \uu\big),&\forall (i,j)\in\EE \label{eq:En_edge_update} \\
    \vv_i^+ &= \phi^v\big(\rho^{e\to v}\big(\{\ee_{ij}^+\}_{j\in\NN_i}\big), \vv_i, \uu\big),&\forall i\in\VV\label{eq:En_node_update}\\
    \xx_i^+ &= \xx_i + \sum\nolimits_{j \in \NN_i} (\xx_i-\xx_j)\phi^x\left(\ee_{ij}^+, \vv_i^+, \vv_j^+,\uu\right),&\forall i\in\VV \label{eq:En_coord_update}\\
    \uu^+ &= \phi^u\big(\rho^{e\to u}(\{\ee_{ij}^+\}_{(i,j)\in\EE}),\rho^{v\to u}(\{\vv_{i}^+\}_{i\in\VV}), \rho^{x\to u}(\{\|\xx_{i}^+-\xx_j^+\|_2^2\}_{(i,j)\in\EE}), \uu\big)
  \end{align}
\end{subequations}
where, similarly as before, $\phi^e:\R^{n_e+ 2n_v+ n_u+1}\to \R^{n_e}$, $\phi^v:\R^{n_e+ n_v+ n_u}\to \R^{n_v}$, $\phi^x:\R^{n_e+ n_v+ n_u}\to \R^{n_x}$, $\phi^u:\R^{n_e+ n_v+n_x+ n_u}\to \R^{n_u}$ are the update functions to be learned and $\rho^{e\to v}, \rho^{e\to u}, \rho^{x\to u}, \rho^{v\to u}$ are aggregation functions. A graphical representation of the process is depicted in Figure~\ref{fig:gnns}.

As the node update has not changed with respect to a standard GNN, but depends on the updated edges $\ee_{ij}^+$, we need to show that the edge update,~\eqref{eq:En_edge_update} is equivariant under $E(n)$. Thanks to Eq.~\eqref{eq:En} we know that $ \|\xx_i-\xx_j\|_2^2$ is equivariant under an $E(n)$ transformation, and we can easily show
\begin{align*}
 \phi^e\big(\ee_{ij}, \vv_i, \vv_j, \|\xx_i-\xx_j\|_2^2, \uu\big) & \to  \phi^e\big(\ee_{ij}, \vv_i, \vv_j, \|Q\xx_i + \zz -Q \xx_j + \zz\|_2^2, \uu\big) \\
&=   \phi^e\big(\ee_{ij}, \vv_i, \vv_j, \|\xx_i - \xx_j \|_2^2, \uu\big) \,.
\end{align*}
As for the coordinate update, Eq.~\eqref{eq:En_coord_update} is $E(n)$ equivariant, since, under an $E(n)$ transformation one has
\begin{align*}
\xx_i  ^+ \to Q  \xx_i  ^+ + \zz &= Q\xx _i + \zz + \sum\nolimits_{j \in \NN_i} (Q\xx_i + \zz- Q\xx _j - \zz )\phi^x(\ee ^+_{ij},\vv ^+_i, \uu) \\
& = Q \left[ \xx _i + \sum\nolimits_{j \in \NN_i} (\xx_i  - \xx _j) \phi^x(\ee ^+_{ij},\vv ^+_i, \uu) \right]+ \zz \\
& = Q \xx_i  ^+ +\zz \,,
\end{align*}
where $\phi^x$ is trivially $E(n)$ equivariant due to the construction of the edge and node updates, Eqs.~\eqref{eq:En_edge_update},\eqref{eq:En_node_update}. Finally, the equivariance of the global update follows naturally.

We point out that the framework presented thus far is quite general and a number of architectures can be translated into the same formalism (as in~\citep{battaglia2018relational}). As an example, consider the case in which one wishes to infer edges in the network. In this case it is possible to start by working with a fully connected graph (\ie, a graph where $\NN_i=\VV\setminus \{i\}$ for all $i$). Then, the summation $ \sum\nolimits_{j \in \NN_i}$ in Eq.~\eqref{eq:En_coord_update} becomes $\sum_{j\neq i}$ and the aggregation in Eq.~\eqref{eq:En_node_update} will automatically generalise.

\section{Conclusion}
In this paper we have outlined a novel deep learning architecture which is equivariant to the $S_n$ and $E(n)$ groups.
The graph network we describe is defined in its most general form, allowing for the model to be applied to a wide range of applications for which graphically structured data is suitable.
We have discussed how including equivariance under the Euclidean group may offer major benefits to deep learning of graphically structured data. Future works will include further developments of the proposed architecture, possibly obtaining equivariance to larger groups and rigourous tests on a wider variety of datasets and tasks.

\bibliographystyle{plainnat}
\bibliography{biblio}

\end{document}